%% file: elsarticle-template.tex
\documentclass[review]{elsarticle}

\usepackage{lineno,hyperref}
\modulolinenumbers[5]

\usepackage{graphicx}
\usepackage{multirow}
\usepackage{amsmath}
\usepackage{booktabs}
\usepackage{subfigure}
\usepackage{bbding}
\usepackage{color}
\usepackage{comment}
\usepackage{pifont}
\newcommand{\cmark}{\ding{51}}%
\newcommand{\xmark}{\ding{55}}%

\journal{Neurocomputing}

\begin{document}

\begin{frontmatter}

\title{Jointly Modeling Hierarchical and Horizontal Features for  Relational Triple Extraction}

\author[jlu]{Zhepei~Wei}
\author[hw]{Yantao~Jia}
\author[jlu]{Yuan~Tian}
\author[uoe]{Mohammad~Javad~Hosseini}
\author[pku]{Sujian~Li}
\author[uoe]{Mark~Steedman}
\author[jlu]{Yi~Chang}
 
\address[jlu]{School of Artificial Intelligence, Jilin University, Changchun, China}
\address[hw]{Huawei Technologies Co., Ltd, Beijing, China}
\address[uoe]{School of Informatics, University of Edinburgh, Edinburgh, Scotland}
\address[pku]{Institute of Computational Linguistics, Peking University, Beijing, China}

\begin{abstract}
Recent works on relational triple extraction have shown the superiority of jointly extracting entities and relations over the pipelined extraction manner. 
However, most existing joint models fail to balance the modeling of entity features and the joint decoding strategy, and thus the interactions between the entity level and triple level are not fully investigated. 
In this work, we first introduce the \emph{hierarchical dependency} and \emph{horizontal commonality} between the two levels, and then propose an entity-enhanced dual tagging framework that enables the triple extraction (TE) task to utilize such interactions with self-learned entity features through an auxiliary entity extraction (EE) task, without breaking the joint decoding of relational triples.
Specifically, we align the EE and TE tasks in a position-wise manner by formulating them as two sequence labeling problems with identical encoder-decoder structure. 
Moreover, the two tasks are organized in a carefully designed parameter sharing setting so that the learned entity features could be naturally shared via multi-task learning. 
Empirical experiments on the NYT benchmark demonstrate the effectiveness of the proposed framework compared to the state-of-the-art methods.
\end{abstract}


\end{frontmatter}


\newpage

\input{tex/introduction}
\input{tex/framework}
\input{tex/experiments}
\input{tex/conclusion}
 \section*{Acknowledgments}
This work is supported by the National Natural Science Foundation of China (No.61976102, No.U19A2065), and Jilin Province Science and Technology Growth Program for Youths (No. 20210508060RQ). 
\section*{References}

\bibliography{HMT}

\end{document}

%% file: tex/introduction.tex
\section{Introduction}
Extracting relational facts from natural language texts is a crucial step towards building large-scale knowledge graph (KG) with entities of different types as nodes and relations among them as edges. 
The facts in KGs are stored in the form of relational triples $(s, r, o)$, in which $s$ and $o$ are the subject and object entities respectively, and $r$ is the relation between them. Typically, the triple extraction (TE) task has been artificially broken down to a pipeline of entity extraction (EE)~\cite{ratinov2009Design} and relation classification (RC) assuming the entity pairs $(s,o)$ in context $x$ are given~\cite{hoffmann2011Knowledge, gormley2015Improved,lin2016Neural}.
However, this separate pipelined setting neglects the interactions between the two steps, and inevitably suffers from the error propagation problem~\cite{zheng2017Joint}. 
To ease the above issues, recent works have explored the task in a joint manner~\cite{li2014Incremental}, which aims to learn relational triples end-to-end, instead of learning entities and relations with two independent models.

Though previous works investigated numerous joint models~\cite{miwa2016End,sun2018Extracting,bekoulis2018Adversarial,gupta2016Table,adel2017Global,miwa2014Modeling}, most of them still follow the pipelined problem formulation  $f(s,o|x) \rightarrow r$, i.e., they achieve joint learning only through parameter sharing between the entity detector and relation classifier, instead of joint decoding~\cite{wang2018Joint}.
In this separate decoding setting,  the relational triples are assembled by pairing the extracted entities to the relation classifier, or by performing table-filling, an exhaustive enumeration of  token pairs to identify triples, which usually suffers from a heavy computational burden. 
Recently, various problem formulations for triple extraction have been proposed, which can be categorized into decompositional (i.e., separate decoding) and non-decompositional (i.e, joint decoding) ones.
Among them are \uppercase\expandafter{\romannumeral1})~$f(r|x)\rightarrow(s,o)$:~modeling subject and object as two arguments of relation~\cite{takanobu2019hierarchical}, \uppercase\expandafter{\romannumeral2})~$f_r(s|x)\rightarrow o$:~modeling relation as function that maps subject to object~\cite{wei2020Novel}, and \uppercase\expandafter{\romannumeral3})~$f(x)\rightarrow (s,r,o)$:~directly modeling the triple without decomposition, which is also referred to as joint decoding~\cite{zheng2017Joint,zeng2018Extracting,zeng2019Learning}. 
Under these novel formulations, such joint models neither need to exhaustively enumerate the token/entity pairs nor have to determine the relation only after all the entities have been recognized as in the traditional pipelined formulation.

Despite their preliminary success, the implementations of these works can hardly balance the modeling of entity features and the joint decoding of triples. 
In fact, they either break the triple extraction into several decompositional modules for a flexible modeling of entity features~\cite{bekoulis2018Adversarial,takanobu2019hierarchical,wei2020Novel}, or compromise on entity features for the sake of a perfect joint decoding strategy~\cite{zheng2017Joint,zeng2018Extracting,zeng2019Learning}, thus leading to a huge performance gap between the entity level and triple level. 
For example, the F1-score for entity extraction could be up to 81.64\% while the performance for triple extraction barely reaches 47.45\% as reported in~\cite{bekoulis2018Adversarial}.

\begin{figure*} [!t]
	\centering
	\includegraphics[width=1\linewidth]{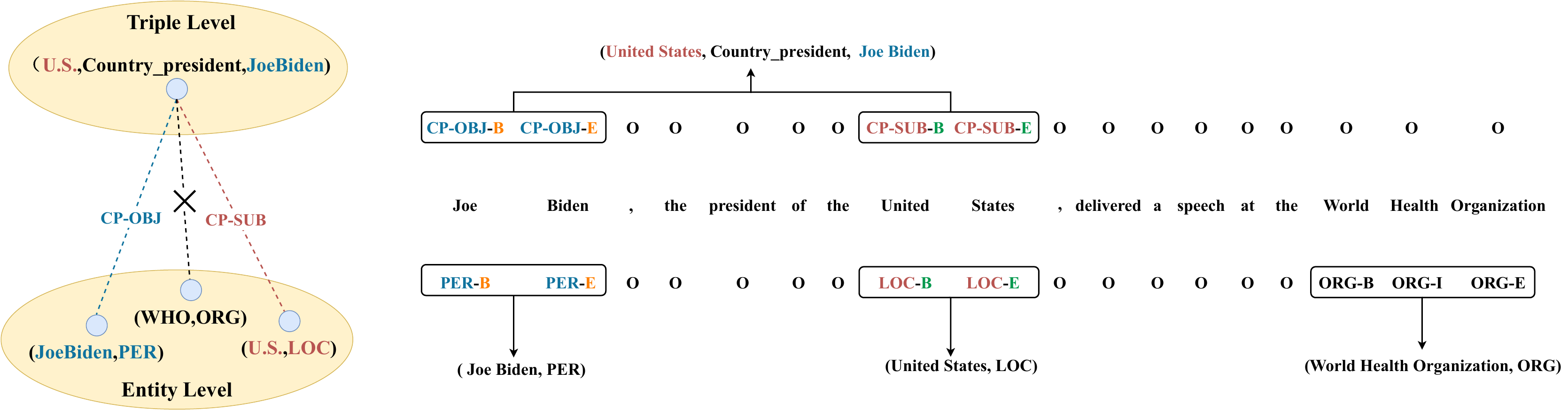}
	\caption{An illustration of hierarchical dependency (left) and horizontal commonality (right) between the entity level and triple level. In this example, ``CP" is the abbreviation for relation ``Country--President". ``SUB" and ``OBJ" are the abbreviations for subject and object entity. ``B", ``I" and ``E" are the abbreviations for  ``Begin", ``Inside" and ``End", representing the boundaries of entity. The design of tagging scheme is detailed in Section~\ref{tag_scheme}.
	}\label{fig:tag}
\end{figure*}

Specifically, there are two-folds interactions between entity level and triple level, namely, the \emph{hierarchical dependency feature} (HDF) and \emph{horizontal commonality feature} (HCF) as illustrated in Fig.~\ref{fig:tag}. 
The HDF indicates that a triple with certain relation type has implicit entity type constraint for candidate subjects and objects. 
For instance, given a relational triple with relation type ``Country-President", i.e., \textit{(Subject, Country-President, Object)}, it is supposed to take an entity with type ``LOC" (instead of ``ORG") as subject entity, and take an entity with type ``PER" as object entity, according to the intrinsic semantic constraints of its relation type. 
Therefore, most existing methods model the HDF by treating the identification of entity and its type as a sub-task of triple extraction under the decompositional perspective. 
Intuitive it may be, this decompositional perspective also limits the exploration of attempts to utilize deeper interactions between entity level and triple level.
In contrast, the emergence of joint decoding strategy brings a fresh non-decompositional perspective to revisit the interaction between the two tasks.
Specifically, it shed light on the inherent horizontal commonality that has rarely been explored before.
By definition, given the input sentence, the EE task is expected to output pairs of entities as well as their types, i.e., (EntityMention, EntityType).
Meanwhile, the TE task aims to obtain triples of two roled entities together with the involved relation, i.e., (Subject, Relation, Object).
Under the non-decompositional perspective, the HCF reveals the commonality that both tasks necessitate identifying the boundaries of entities from the context.
Hence, we argue that the position between the two tasks could be more than simply subordinate but also reciprocal if the HCF is appropriately modeled.
Unfortunately, this basic fact has been long overlooked in previous works.

Such hierarchical dependency and horizontal commonality straightforwardly motivates an intuitive idea of employing multi-task learning for relational triple extraction, where TE as the main task and EE as the auxiliary task.
However, this brings significant difficulties for the practical design of two tasks.
First, the complex hierarchical and horizontal interactions directly challenge the traditional multi-task setting that adopts soft or hard parameter sharing. 
Second, despite there being various extractive and generative models for the two tasks, the horizontal commonality requires the two tasks to be strictly aligned under the non-decompositional perspective, which inevitably limits the options of problem formulation.
Therefore, the core idea of jointly modeling the hierarchical and horizontal features for relational triple extraction lies in an easy-to-share multi-task setting with well-aligned problem formulations of EE and TE tasks.

In this work, we implement the above idea in an entity-enhanced dual tagging framework.
Specifically, the EE and TE tasks are formulated as two aligned sequence labeling problems with identical structure and carefully designed tagging schemes.
Unlike the soft or hard parameter sharing, each task has its own module with independent parameters but with partial overlap in an customized setting, as illustrated in Fig.~\ref{fig:sharing}.
In this way, the HDF and HCF can be naturally captured via multi-task learning, making the entity enhanced by both type constraint and boundary information.
More importantly, the proposed dual tagging framework achieves the joint modeling of such interactions without breaking the joint decoding strategy, which makes our framework essentially different from existing decompositional and non-decompositional multi-task learning models~\cite{miwa2016End,takanobu2019hierarchical,wei2020Novel,zeng2020copymtl}.
We compare our framework with numerous state-of-the-art baselines, including six decompositional models and two non-decompositional models. 
Empirical experiments show the superiority of the proposed framework, highlighting the importance of joint modeling the hierarchical and horizontal  features under the non-decompositional perspective for relational triple extraction.
The main contributions of this work are as follows:
\begin{itemize}
	\item [1.] We introduce the hierarchical dependency and horizontal commonality  between entity level and triple level in the task of relational triple extraction. To the best of our knowledge, we are the first to collectively investigate such interactions under the non-decompositional perspective.
	\item [2.] We implement an entity-enhanced dual tagging framework where EE and TE tasks are formulated as two aligned sequence labeling problems and optimized via multi-task learning. It naturally exploits the above interactions while keeping the joint decoding strategy for triple extraction.
	\item [3.] Our method substantially outperforms the state-of-the-art models on the NYT benchmark with 3.3\% improvements in terms of F1-score. Additionally, we further explore the utility of the proposed dual tagging setting in detail through extensive analysis.
\end{itemize}

%% file: tex/framework.tex
\section{Entity-enhanced Dual Tagging Model}\label{sec:framework}
\begin{figure} [!t]
	\centering
	\subfigure[Hard]{
		\includegraphics[width=0.316\linewidth]{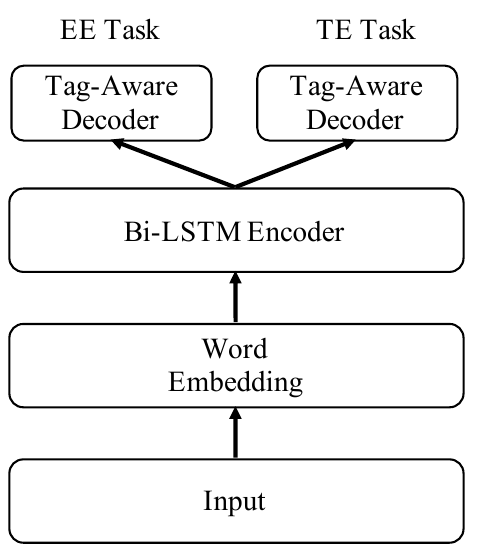}}\label{fig:soft_sharing}
	\subfigure[Soft]{
		\includegraphics[width=0.316\linewidth]{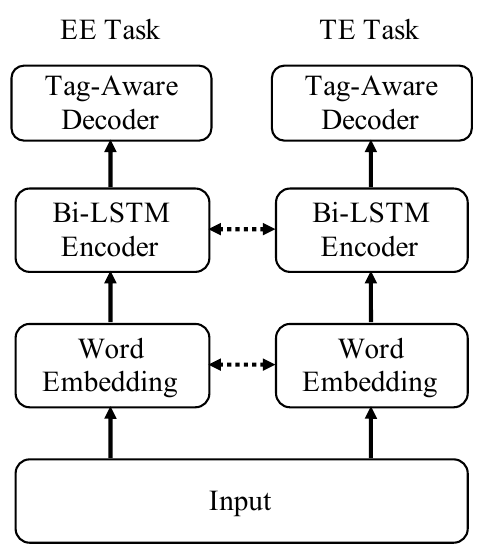}}
	\subfigure[Ours]{
		\includegraphics[width=0.316\linewidth]{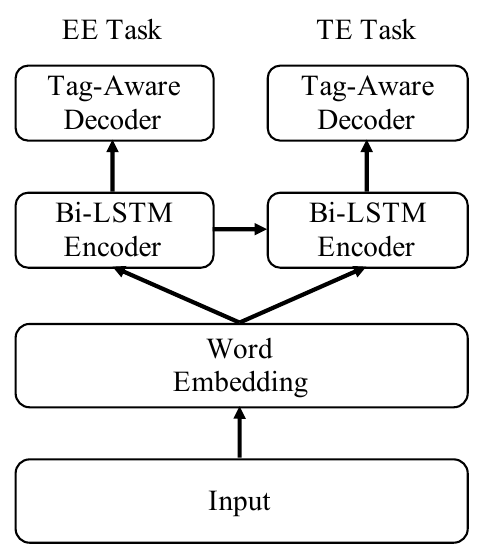}}
	\caption{(a) illustrates the hard parameter sharing; (b) illustrates the soft parameter sharing; (c) illustrates our customized parameter sharing. Solid~line represents the data flow. Dashed line represents the parameter constraint, which typically with L2 regularization. For a more in-depth discussion of parameter sharing in MTL, we refer readers to \cite{ruder2017overview}.}
	\label{fig:sharing}
\end{figure}

In this section, we elaborate on the design of the proposed dual tagging framework.
To align the two tasks, each of which is implemented as a sequence labeling module with the same encoder-decoder structure.
Specifically, we adopt a bidirectional \emph{long short-term memory} (Bi-LSTM) encoder and a tag-aware LSTM decoder.
Typically, multi-task learning (MTL) in deep learning model is done with either hard or soft parameter sharing of hidden layers\cite{ruder2017overview}.
However, neither of them could model the semi-hierarchy and semi-parallel nature between the EE and TE tasks.
Hence, we introduce a customized method targeted on such complex interactions, which smoothly captures the HDF and HCF.
The difference among these methods is illustrated in~Fig.\ref{fig:sharing}.

\subsection{Tagging Scheme}\label{tag_scheme}
Before diving into the model structure, we first introduce the design of tagging schemes of the two tasks.
As we align the EE and TE tasks as two sequence labeling problem, the goal of each task is to assign a pre-defined tag to each token of the input sentence and obtain the extracted results according to the output tag sequence.
For EE task, let $ \mathcal{E}$ denote the set of entity types, such as \{``PER", ``LOC",  ``ORG", ``MISC"\} (person, location, organization, miscellaneous).
Let $\mathcal{P}$ denote the set of position indicators, such as \{``B",``I", ``O", ``E", ``S"\} (Begin, Inside, Outside, End, Single) or \{``B",``I", ``O"\}.
The pre-defined tag set for EE task is $\mathcal{T}_{EE} = \mathcal{E} \times \mathcal{P'}~\cup $ \{``O"\} $=\{(e,p)|e \in \mathcal{E}, p \in \mathcal{P'}\}~\cup $ \{``O"\}, where $\mathcal{P'} = \mathcal{P}~\setminus$ \{``O"\}. 
Hence, the size of EE tag set is $|\mathcal{E}|*|\mathcal{P'}|+1$, where $|\mathcal{E}|$ is the number of pre-defined entity types, $|\mathcal{P'}|$ is subject to the position indicators, for instance, $|\mathcal{P'}|=4$ if the ``BIOES" indicators are adopted.
Among the tag set, tag ``O" means the corresponding token is irrelevant to any concerned entity.
The rest tags are composed of two parts: \emph{entity type} and \emph{position indicator}.
Specifically, the \emph{entity type} part reveals that the corresponding token is part of an concerned entity.
The \emph{position indicator} part reveals which part of the concerned entity the corresponding token belongs to.

Likewise, for TE task,  let $ \mathcal{R}$ denote the set of relation types, $\mathcal{P}$ denote the set of position indicators, and $ \mathcal{S}$ denote the set of semantic role, i.e, \{``SUB",``OBJ"\} (Subject, Object).
The pre-defined tag set for EE task is $\mathcal{T}_{TE} = \mathcal{R} \times\mathcal{ S} \times \mathcal{P'}~\cup$ \{``O"\} $=\{(r,s,p)|r \in \mathcal{R}, s \in \mathcal{S}, p \in \mathcal{P'}\} ~\cup $ \{``O"\}, where $\mathcal{P'} = \mathcal{P}~\setminus$ \{``O"\}. 
Specifically, all tags except ``O" are  composed of three parts: \emph{relation type}, \emph{semantic role} and \emph{position indicator}.
The \emph{relation type} part reveals that the corresponding token is part of an concerned triple.
The  \emph{semantic role} part reveals the semantic role of the token in the concerned triple. In other words, the corresponding token is either part of a subject entity or an object entity. 
The \emph{position indicator} part is similar to that of EE task. 

Here we take the TE tags for example and show how we obtain the relational triples based on the predicted tags.
As shown in Fig.~\ref{fig:tag}, ``Joe" is the first word of entity ``Joe Biden", which is involved in the relation ``Country--President" as the object entity. Hence, the tag of ``Joe" is ``CP-OBJ-B".
Meanwhile, the first word ``United" in ``United States" is tagged as ``CP-SUB-B" because ``United States" is involved in the relation ``Country--President" as the subject entity.
To obtain the final relational triples, we combine the entities with the same relation type according to their tags.
Now we know ``Joe Biden" and ``United States" share the same relation type ``Country--President" as well as their semantic roles in the relation.
Therefore, the final extracted relational triple is (United States, Country--President, Joe Biden).

\subsection{Embedding Layer}
Word embedding~\cite{hinton1986Learning} is proposed to convert the words into vector representations that capture the syntactic and semantic meanings of words. 
An input sentence $s=\{\text{x}_1, \text{x}_2, ..., \text{x}_n\}$, where $\text{x}_i$ is the i-th word, can be represented by a sequence of vectors $\{w_1, w_2, ..., w_n\}$ and then fed into neural networks. In this work, we use the pre-trained Glove~\cite{pennington2014Glove} word embeddings.
\begin{figure} [!t]
	\centering
	\includegraphics[width=1\linewidth]{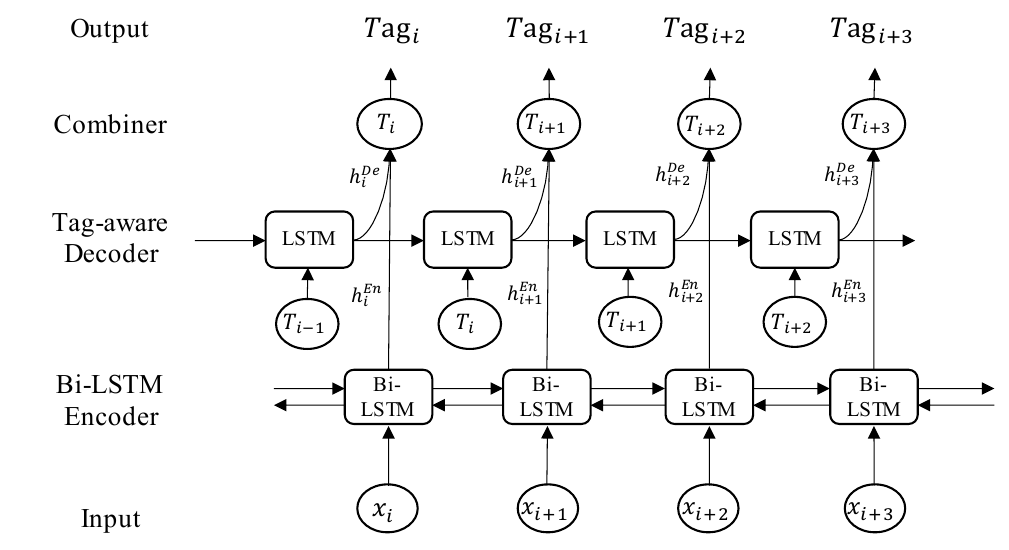}
	\caption{The Encoder-Decoder Structure.
		``En" represents encoder and ``De" represents decoder. ``Tag" represents the EE/TE tag.}\label{fig:en-de}
\end{figure}
\subsection{Encoder-Decoder Module}
Both the EE and TE tasks are aligned in the same form of sequence labeling, with an identical encoder-decoder structure as shown in Fig.\ref{fig:en-de}.
For clarity, we avoid repeating the model description for both tasks but generally introduce the encoder-decoder structure from a unified perspective.
The task-specific differences are detailed in the following part as well. 

\subsubsection{Bi-LSTM Encoder} 
The bidirectional \emph{long short-term memory} (Bi-LSTM) has been proved~\cite{lample2016Neural} to be effective in capturing semantic information for each word in the input sentence by processing the sequence in both directions with two parallel LSTM layers.
In this work, we use a popular LSTM variant~\cite{gers2000recurrent} as the encoder.
The detailed operations are as follows:
\begin{align}
	&f_t = \sigma(W_{f_x}x_t + W_{f_h}h_{t-1} + W_{f_c}c_{t-1} + b_f)\nonumber \\
	&i_t = \sigma(W_{i_x}x_t + W_{i_h}h_{t-1} + W_{i_c}c_{t-1} + b_i)\nonumber\\
	&c_t = f_tc_{t-1} + i_t\tanh(W_{c_x}x_t + W_{c_h}h_{t-1} + b_c\nonumber)\\
	&o_t = \sigma(W_{o_x}x_t + W_{o_h}h_{t-1} + W_{o_c}c_t + b_o)\nonumber\\
	&h_t = o_t\tanh(c_t)
\end{align}
where at time step $t$ $(1\leq t\leq n)$,  $x_t, h_t$ and $c_t$ are input vector, hidden state vector  and cell state vector respectively, $f_t, i_t$ and $o_t$ are forget gate, input gate and output gate respectively, and $b_{(.)}$ is the bias weight.
For each input $x_t$ in a sequence of length $n$, the forward LSTM encodes $x_t$ by considering the contextual information from $x_1$ to $x_t$, and the encoded vector is denoted as $\overrightarrow{h_t}$.
Similarly, the backward LSTM encodes $x_t$ based on the contextual information from $x_n$ to $x_t$, and the encoded vector is denoted as $\overleftarrow{h_t}$.
We then concatenate $\overrightarrow{h_t}$ and $\overleftarrow{h_t}$ to represent the t-th word of the input sequence and denote it as $h_t^{En} = [\overrightarrow{h_t}, \overleftarrow{h_t}]$. 

According to our customized parameter sharing setting,  the input of the EE encoder  is the word embedding vector $w_t$ of t-th word $\text{x}_t$, i.e., $x_t = w_t$.
For TE task, the input of the encoder is the concatenation of embedding vector $w_t$ and encoded vector $h_t^{En}$ from EE task, i.e., $x_t = [w_t, h_t^{En}]$.

\subsubsection{Tag-aware LSTM Decoder} We also design a tag-aware LSTM decoder to predict the tag sequences.
To explicitly model the interactions between tags, we fuse the tag information into the input of LSTM at each time step $t$. The detailed operations are as follows: 
\begin{align}
	&f_t = \sigma(W_{f_T}T_{t-1} + W_{f_h}h_{t-1} + b_f)\nonumber\\
	&i_t = \sigma(W_{i_T}T_{t-1} + W_{i_h}h_{t-1} + b_i)\nonumber\\
	&c_t = f_tc_{t-1} + i_t\tanh(W_{c_T}T_{t-1} + W_{c_h}h_{t-1} + b_c)\nonumber\\
	&o_t = \sigma(W_{o_T}T_{t-1} + W_{o_h}h_{t-1} + b_o)\nonumber\\
	&h_t = o_t\tanh(c_t)
\end{align}
where $T_t$ is the predicted tag vector of the t-th word $\text{x}_t$. The decoder hidden state vector at time step $t$ is denoted as $h_t^{De}$, which is concatenated with encoded vector $h_t^{En}$ for computing  $T_t$:
\begin{equation}
	T_t = W_T[h_t^{En}, h_t^{De}] + b_T
	\label{formula:T}
\end{equation}
We can then compute the tag probability for the t-th word $\text{x}_t$ in sentence $s_j$ based on the predicted tag vector $T_t$:
\begin{equation}
	y_t = W_yT_t + b_y
\end{equation}
\begin{equation}
	p(y_t^i|s_j,\theta) = \frac{\exp(y_t^i)}{\sum_{k=1}^{N_t}\exp(y_t^k)}
	\label{formula:prob}
\end{equation}
where $p(y_t^i|s_j,\theta)$ is the probability of assigning the i-th tag to the t-th word $\text{x}_t$ in sentence $s_j$, $\theta$ is the model parameter and $N_t$ is the number of tags. 

\subsection{Multi-task Objective Function}
Since the  EE and TE tasks are formulated as two aligned sequence labeling problems, both of them hold the same form of objective function:
\begin{align}
	J(\theta)_{(.)} = \max\sum_{j=1}^{|D|}\sum_{t=1}^{L_j}(\log p(y_t^{G_t}|s_j,\theta))
\end{align}%
where $|D|$ is the size of training set, $L_j$ is the length of sentence $s_j$, $G_t$ is the gold standard tag of t-th word $\text{x}_t$ in $s_j$, and $p(y_t^{G_t}|s_j,\theta)$ is the probability of assigning $G_t$ to $\text{x}_t$, which is defined in Equation~\ref{formula:prob}.
\par The final multi-task objective function of the dual tagging framework is defined as follows:
\begin{align}
	J(\theta) = \alpha J_{EE}(\theta) + (1 - \alpha) J_{TE}(\theta) \label{formula:loss}
\end{align}%
where $\alpha \in (0, 1)$ is the bias hyper-parameter. We train the model by maximizing the log likelihood $J(\theta)$ through stochastic gradient descent over shuffled mini-batches and the optimization algorithm is RMSprop~\cite{tieleman2012Lecture}.

%% file: tex/experiments.tex
\section{Experiments} 
In this section, we conduct extensive experiments and in-depth analysis, aiming to answer the following research questions:
\begin{itemize}
	\item \textbf{Q1}: Does our method outperform the state-of-the-art baselines? (section \ref{sec:results})\label{exp:q1}
	\item \textbf{Q2}: Is the dual tagging setting beneficial to relational triple extraction? If yes, which part of triple does it contribute to? (section \ref{sec:ablation_study})\label{exp:q2}
	\item \textbf{Q3}: What is the effect of different tagging schemes? How important is EE task to TE task? (section \ref{sec:hyper-parameter_study})\label{exp:q3}
	\item \textbf{Q4}: In which case does our method work well or fail? (section \ref{sec:case_study})\label{exp:q4}
\end{itemize}
The experimental setting,  results, analysis and all compared baselines are also detailed in the rest of this section.

\subsection{Experimental Setting}
\subsubsection{Dataset and Evaluation Metrics} 
Following~\cite{takanobu2019hierarchical}, we use the public New York Times (NYT) corpus to evaluate the proposed  framework, in which the training set is produced by distant supervision~\cite{riedel2010Modeling} while the test set is manually annotated~\cite{hoffmann2011Knowledge}. 
Specifically, it contains 62,648 sentences for training and 369 sentences for test.
We also create a validation set by randomly sampling 0.5\% data from the training set as inline with previous work~\cite{takanobu2019hierarchical}. 

We report the standard micro Precision, Recall, and F1-score to evaluate our model.
An extracted relational triple is regarded as correct only if the relation type and the heads of both subject and object are all correct, for a fair comparison with baseline models~\cite{zeng2018Extracting,takanobu2019hierarchical,wei2020Novel}.

\subsubsection{Implementation Details}
We finetune the hyper-parameters on validation set.
For embedding layer, the dimension of word embeddings is 300.
For the encoder-decoder module, the hidden size of LSTM cell is 300 in both Bi-LSTM encoder and tag-aware LSTM decoder.
The dimension of tag vector $T$ in Equation~\ref{formula:T} is 300.
The hyper-parameter $\alpha$ in Equation~\ref{formula:loss} is 0.9.
We adopt ``BIOES" based scheme for the dual tagging framework.
We also adopt dropout and mini-batch mechanism in our model where the dropout rate is 0.5 and the batch size is 256.
\renewcommand{\arraystretch}{1} 
\begin{table*}[!t] 
	\centering 
	\caption{Results of different methods on relational triple extraction. \textit{ExtraRes} indicates the method exploits additional resources besides the input raw sentence, including dependency tree path, part-of-speech tag and other information such as entity types and descriptions, etc. Notation ``$\dagger$" represents the results are copied from~\cite{takanobu2019hierarchical}, ``$\ddagger$" represents our implementation.}
	\resizebox{1\linewidth}{!}{
		\begin{tabular}{llcccccc} 
			\toprule
			Method&\textit{ProbForm}&\textit{HDF}&\textit{HCF}&\textit{ExtraRes}&\textit{Precision}&\textit{Recall}&\textit{F1-score}\\
			\midrule 
			MultiR$^{\dagger}$
			~\cite{hoffmann2011Knowledge}
			&$f(s,o|x) \rightarrow r$&\xmark&\xmark&\cmark&0.328&0.306&0.317\\
			FCM$^{\dagger}$ 
			~\cite{gormley2015Improved}
			&$f(s,o|x) \rightarrow r$&\cmark&\xmark&\xmark&0.432&0.294&0.350\\
			CopyR$^{\dagger}$
			~\cite{zeng2018Extracting}
			&$f(x) \rightarrow (s,r,o)$&\xmark&\xmark&\xmark&0.347&0.534&0.421\\
			CoType$^{\dagger}$
			~\cite{ren2017Cotype}
			&$f(s,o|x) \rightarrow r$&\cmark&\xmark&\cmark&0.486&0.386&0.430\\ 
			NovelTag$^{\dagger}$
			~\cite{zheng2017Joint}
			&$f(x) \rightarrow (s,r,o)$&\xmark&\xmark&\xmark&0.469&0.489&0.479\\ 
			SPTree$^{\dagger}$
			~\cite{miwa2016End}
			&$f(s,o|x) \rightarrow r$&\cmark&\xmark&\cmark&0.522&0.541&0.531\\
			HRL$^{\dagger}$
			~\cite{takanobu2019hierarchical}
			&$f(r|x) \rightarrow (s,o)$&\xmark&\xmark&\xmark&0.538&0.538&0.538\\
			CasRel$^{\ddagger}$
			~\cite{wei2020Novel}
			&$f_r(s|x) \rightarrow o$&\xmark&\xmark&\xmark&0.538&\textbf{0.597}&0.566\\
			\midrule
			Ours$^{\ddagger}$ (BIO)
			&$f(x) \rightarrow (s,r,o)$&\cmark&\cmark&\xmark&0.648 $\pm$ 0.016&0.540 $\pm$ 0.014&0.589 $\pm$ 0.012\\
			Ours$^{\ddagger}$ (BIOES)
			&$f(x) \rightarrow (s,r,o)$&\cmark&\cmark&\xmark&\textbf{0.662} $\pm$ 0.012& 0.546 $\pm$ 0.009&\textbf{0.599} $\pm$ 0.009\\
			\bottomrule 
		\end{tabular} 
	}
	\label{tab:res} 
\end{table*}

\subsection{Experimental Results}
\subsubsection{Baselines}
We compare the proposed framework with several representative advanced triple extraction models, including a traditional pipelined model (FCM) and some recent joint models that adopts various decompositional (MultiR, CoType, SPTree, HRL and CasRel) and non-decompositional (NovelTag, CopyR) problem formulations.
As the proposed dual tagging framework is not limited to any specific tagging scheme, we investigate the performance with both ``BIO'' and ``BIOES'' based schemes.
To validate the robustness of our method, we report the averaged results of five runs and the corresponding standard deviation.

\subsubsection{Main Results} \label{sec:results}Table~\ref{tab:res} shows the results of different methods on relational triple extraction.
In response to the research question \textbf{Q1}, we can find that the proposed model substantially outperforms all the baselines in terms of F1-score.
More precisely, it achieves 3.3\% improvement over the best joint model CasRel~\cite{wei2020Novel} on the NYT benchmark, demonstrating the superiority of the dual tagging framework.

From the perspective of problem formulation, it can be observed  that models with more advanced forms mostly perform better than those with traditional form of $f(s,o|x) \rightarrow r$. Though there are a few exceptions like SPTree that achieves a slightly higher performance, it requires more linguistic resources (e.g., POS tags, chunks and syntactic parsing trees).
It is also worth noting that CasRel achieves the best performance in recall, this is because it is designed to extract as many triples as possible by exhaustively considering all the relations altogether, which consequently result in a comparatively lower performance in precision.
In contrast, the precision of our framework is much higher than all the other models.
We attribute this to the dual tagging setting that enables the framework to jointly model the  HDF and HCF between the entity level and triple level, which is of great help for triple extraction in enhancing the identification of entity boundary and its role within the triple.
\begin{figure} [!t]
	\centering
	\subfigure[Sub-Rel-Obj]{
		\includegraphics[width=0.41\linewidth]{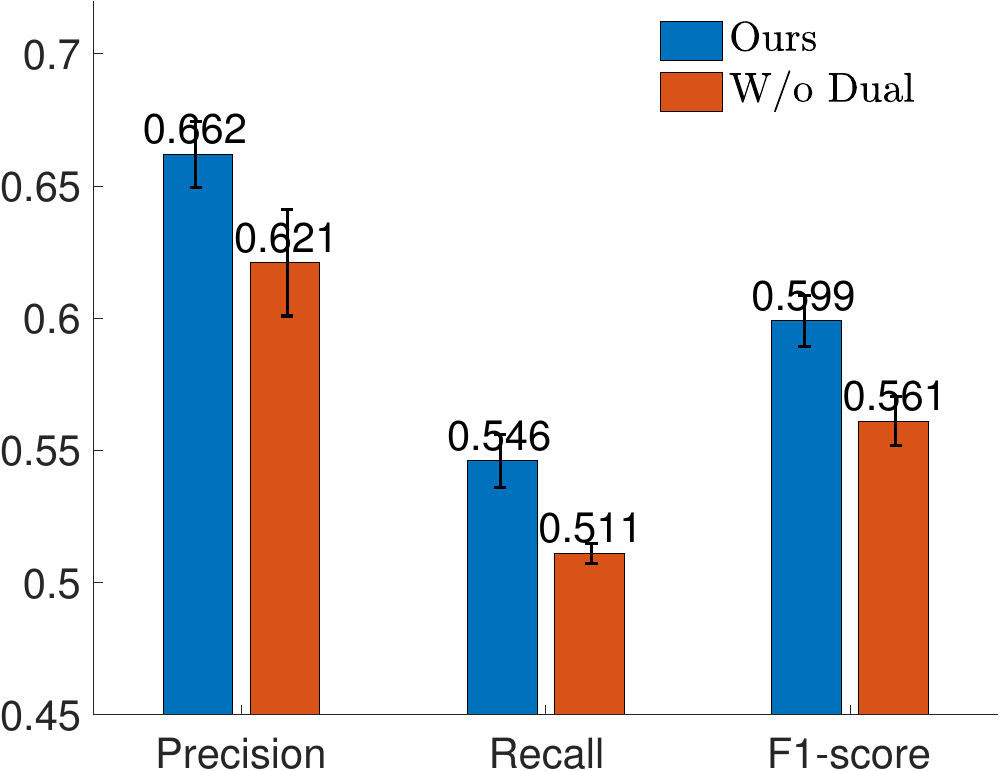}}
	\subfigure[Sub-Rel]{
		\includegraphics[width=0.41\linewidth]{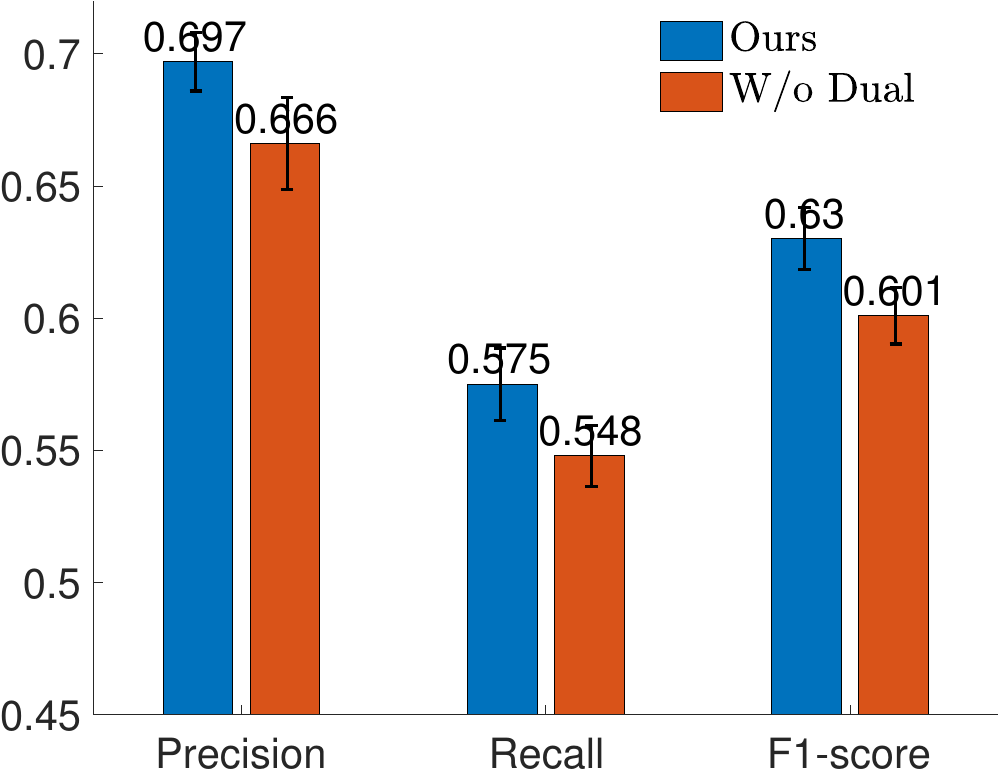}}
	\subfigure[Rel-Obj]{
		\includegraphics[width=0.41\linewidth]{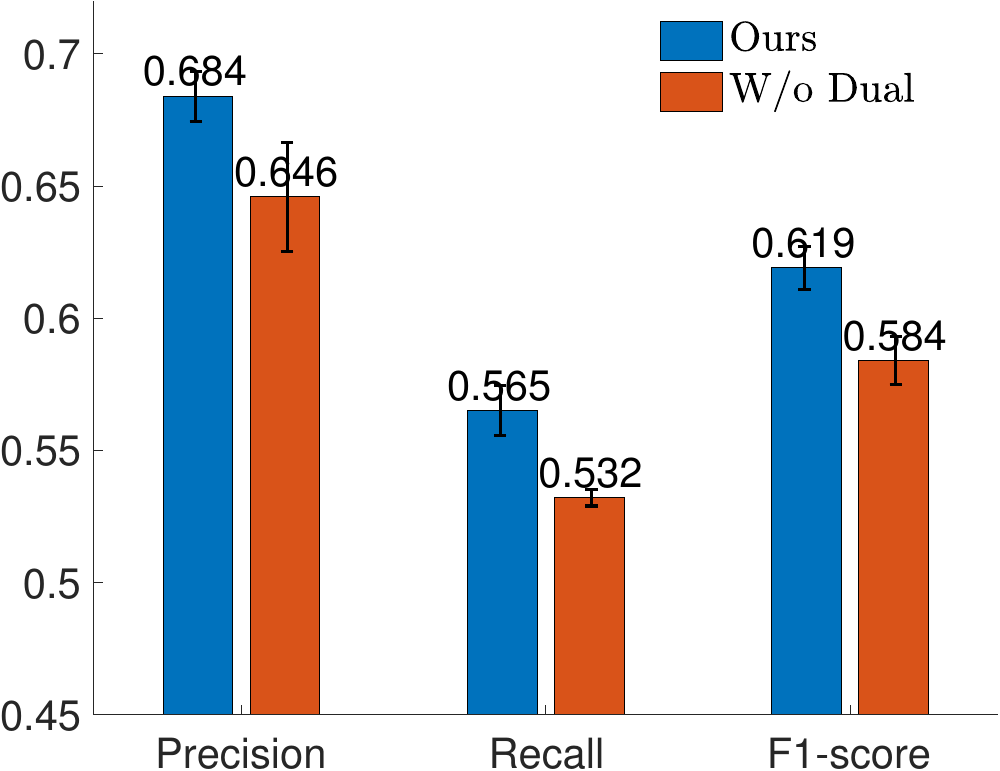}}
	\subfigure[Sub-Obj]{
		\includegraphics[width=0.41\linewidth]{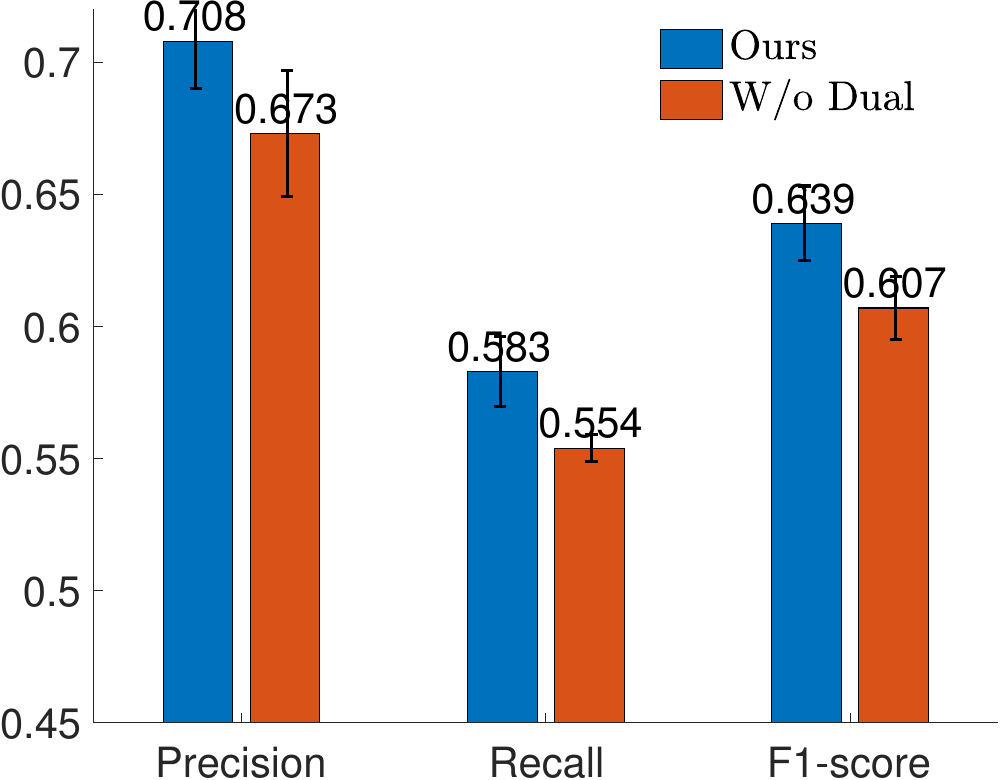}}
	\caption{Results on extracting different elements of relational triple. (a)~evaluating on the complete triple (s, r, o); (b)~evaluating on the subject and relation part (s, r, $\ast$); (c)~evaluating on the relation and object part ($\ast$, r, o); (d)~evaluating on the subject and object part (s, $\ast$, o). Error bars show $\pm$1 standard deviation around the average of 5 runs.}
	\label{fig:dual}
\end{figure}
\subsubsection{Ablation Study} \label{sec:ablation_study}
Regarding to the research question \textbf{Q2}, we conduct a set of ablation studies focusing on the dual tagging setting and make an exhaustive comparison among different elements of the relational triple, as presented in Fig.~\ref{fig:dual}.

To investigate the effectiveness of the dual tagging setting, we test it by removing the EE module from our framework (denoted as ``W/o Dual") and then re-evaluate its performance for five runs. 
Specifically, we determine the scores by cheeking if the relation type and the heads of two entities of the extracted triple (s, r, o) are all consistent with the ground truth label.
According to Fig.~\ref{fig:dual} (a), we can observe that the performance decreases drastically with about 6.2\% and 6.4\% relative loss in precision and recall scores, respectively.
The performance drops between the two setups demonstrate the importance of introducing the dual tagging setting.

Moreover, to make sure which part of triple it mainly contributes to, we further explore the performance on extracting three fine-grained elements of triple as shown in Fig.~\ref{fig:dual} (b) to~\ref{fig:dual}(d).
For example, Sub-Obj represents that we only focus on the subject and object part of the extracted triples.
More precisely, an element (s, $\ast$, o) is determined as correct if the subject and object entity in the extracted triple (s, r, o) are both correct, disregarding of the correctness of its relation type.
The same goes for Sub-Rel and Rel-Obj.
Generally, it can be observed that the dual tagging setting contributes to all three elements of the triple, indicating its utility in our framework.
The exciting finding is the new insight behind the performance changes among these elements when individually looking from the perspective of the two setups.
Under the dual tagging setting, there is a gap (from 0.619 to 0.599) between ($\ast$,~r,~o) and (s,~r,~o), indicating that wrongly identifying the subject leads to about 3.2\% relative loss in performance.
Likewise, the relative loss for misidentifying relation and object approximately reaches 6.2\% and 4.9\%.
In contrast, after removing the dual tagging, the relative loss for misidentifying the subject, relation, and object respectively raise to  3.9\%,  7.5\%, and 6.6\%.
Though the relative loss increases for all elements, it clearly demonstrates that the introduced dual tagging setting contributes the most to the identification of the object entity, followed by relation, and subject entity at the end.
\subsubsection{Hyper-Parameter Study}\label{sec:hyper-parameter_study}
\begin{figure} [!h]
	\centering
	\includegraphics[width=0.6\linewidth]{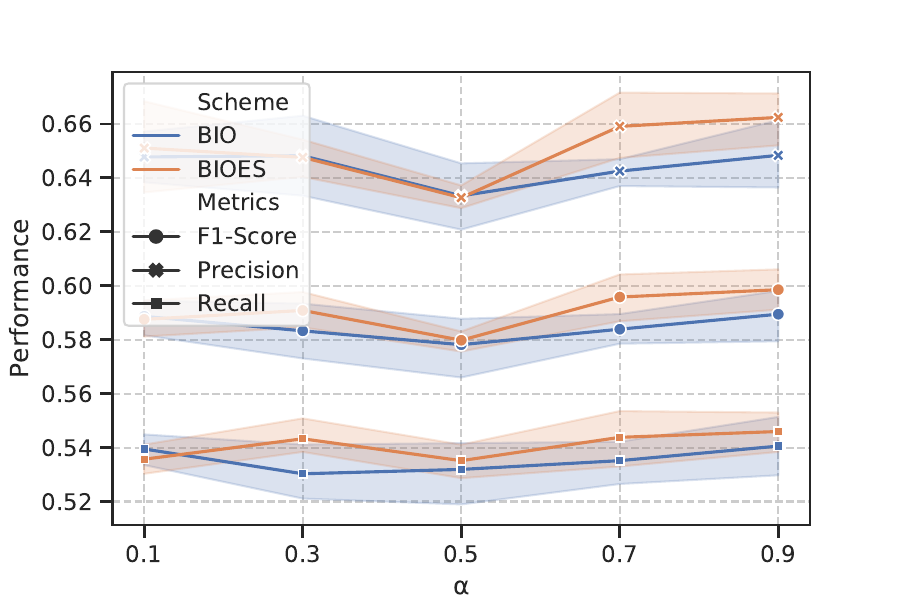}
	\caption{Hyper-Parameter study on bias $\alpha$ and tagging scheme. The colored band for each tagging scheme indicates $\pm$1 standard deviation around the average of 5 runs.}\label{fig:hyper_study}
\end{figure}
For the research question \textbf{Q3}, we conduct a comprehensive hyper-parameter study on the bias between the two tasks and their tagging schemes.
The results are shown in Fig.~\ref{fig:hyper_study}.
Specifically, we jointly consider the combination of  two tagging schemes and five different values of bias $\alpha$ from 0.1 to 0.9 to investigate how important is EE task to TE task, and try to figure out how different tagging schemes affect the interactions between the two tasks.
Generally, it is not surprising to find that the ``BIOES" based scheme performs better than the ``BIO" based one, since it adopts extra tokens to model the boundary information as previous work suggests~\cite{yang2018Design}.
In addition, the standard deviation of ``BIOES" based scheme is consistently lower than that of the ``BIO" based scheme in all three metrics, indicating its superiority in the relational triple extraction task. 

On the other hand, it can be observed that the performance is not always positively related to the bias between the two tasks, though EE is designed to serve as an auxiliary task of TE according to the dual tagging setting.
Instead, it presents a fluctuating trend with the increasing value of~$\alpha$.
It's also surprising to find that the performance reaches the bottom when $\alpha = 0.5$, which means the framework treats the two tasks equally.
Nevertheless, the standard deviation at this point is much lower than that of any other values, indicating its advantage in stability due to the moderate setting. 
Interestingly, though the performance fluctuation is not quite significant among different values of $\alpha$, it is somehow unexpected that $\alpha$ = 0.9 leads to the best performance, which strongly implies the importance of entity features for relational triple extraction.

\begin{table*}[!t]
	\caption{Case study. ``Ground Truth" represents the gold standard tags of the given sentence and ``Predicted" represents the output of our framework. ``PN" is the abbreviation for ``Person--Nationality". ``AC" is the abbreviation for ``Administrative\_division--Country" and ``CA" is the abbreviation for ``Country--Administrative\_division". Note that the relation is directed, ``AC" and ``CA" are actually equivalent. }
	\resizebox{1\linewidth}{!}{
		\begin{tabular}{p{1.8cm}p{1.6cm}p{13cm}}
			\toprule
			Example 1&Ground Truth: &	At the center of this manufactured maelstrom is the preternaturally beauteous figure of $\textbf{[Shilpa Shetty]}_{\color{red} PN-SUB}$ , 31 , a Bollywood movie star from $\textbf{[India]}_{\color{red} PN-OBJ}$ whose treatment by British contestants in the so-called reality show on television here has provoked more than 16,000 viewers to complain to regulators that she is the victim of racist bullying . \\
			&Predicted: &	At the center of this manufactured maelstrom is the preternaturally beauteous figure of $\textbf{[Shilpa Shetty]}_{\color{red} PN-SUB}$ , 31 , a Bollywood movie star from $\textbf{[India]}_{\color{red} PN-OBJ}$ whose treatment by British contestants in the so-called reality show on television here has provoked more than 16,000 viewers to complain to regulators that she is the victim of racist bullying . \\
			\midrule
			Example 2&Ground Truth: &Homage to $\textbf{[Cambodia]}_{\color{blue} AC-OBJ}$ was performed at Chaktomuk Conference Hall in $\textbf{[Phnom Penh]}_{\color{blue} AC-SUB}$ on Oct. 21 , attended by the king . \\
			&Predicted: &Homage to $\textbf{[Cambodia]}_{\color[rgb]{1,0.5,0} CA-SUB}$ was performed at Chaktomuk Conference Hall in $\textbf{[Phnom Penh]}_{\color[rgb]{1,0.5,0} CA-OBJ}$ on Oct. 21 , attended by the king . \\
			\bottomrule
		\end{tabular}
	}
	\label{tab:case}
\end{table*}
\subsubsection{Case Study} \label{sec:case_study}We further study the performance of our method in some real cases to explore the research question \textbf{Q4}.
Table~\ref{tab:case} presents two examples demonstrating the pros and cons.
Specifically, the first example shows that our model can exactly extract relational triple from a long sentence with complex context.
The second example reveals the flaw of our model that it fails to cope with the reversed relations.
Having said that, ``CA" and ``AC" actually express the same relation between two entities but with the opposite direction.
Therefore, if we tolerantly re-evaluate the reversed triples as correct with simple postprocessing, the F1-score could be further boosted from 0.599 to 0.606.
In the future, we shall investigate more relation-sensitive mechanisms to improve the identification of directed relations.

%% file: tex/conclusion.tex
\section{Conclusion}
In this paper, we study on the  relational triple extraction from the non-decompositional perspective.
Specifically, we first introduce the hierarchical dependency and horizontal commonality between entity level and triple level, and then implement a carefully designed dual tagging framework for triple extraction. 
It collectively exploits such interactions on top of the joint decoding strategy via multi-task leaning with a customized  parameter sharing.
Experimental results show that our model substantially outperforms the state-of-the-art baselines on the New York Times (NYT) benchmark. 